\documentclass[runningheads,a4paper]{llncs}
\usepackage{llncsdoc}
\usepackage[export]{adjustbox}
\date{}

\begin{document}
%
%
\pagestyle{empty}

\title{Arithmetic~Word~Problem~Solver using Frame~Identification}
\author{Pruthwik Mishra\and Litton J Kurisinkel \and Dipti Misra Sharma} \institute{KCIS, LTRC, IIIT-Hyderabad\\\email{\{pruthwik.mishra,litton.jkurisinkel\}@research.iiit.ac.in, dipti@iiit.ac.in}}
\maketitle              

\begin{abstract}
Automatic Word problem solving has always posed a great challenge for the NLP community. Usually a word problem is a narrative comprising of a few sentences and a question is asked about a quantity referred in the sentences. Solving word problem involves reasoning across sentences, identification of operations, their order, relevant quantities and discarding irrelevant quantities. In this paper, we present a novel approach for automatic arithmetic word problem solving. Our approach starts with frame identification. Each frame can either be classified as a state or an action frame. The frame identification is dependent on the verb in a sentence. Every frame is unique and is identified by its slots. The slots are filled using dependency parsed output of a sentence. The slots are entity holder, entity, quantity of the entity, recipient, additional information like place, time. The slots and frames helps to identify the type of question asked and the entity referred. Action frames act on state frame(s) which causes a change in quantities of the state frames. The frames are then used to build a graph where any change in quantities can be propagated to the neighboring nodes. Most of the current solvers can only answer questions related to the quantity, while our system can answer different kinds of questions like `who', `what' other than the quantity related questions `how many'.  

There are three major contributions of this paper.
1. Frame Annotated Corpus (with a frame annotation tool)
2. Frame Identification Module
3. A new easily understandable Framework for word problem solving
\end{abstract}
\section{Introduction}
\label{Intro}
In this paper, we developed frame-based Word Problem Solver that solves mathematical problems posed in natural language. The biggest challenge in solving the word problems is the natural language understanding. We present an offline Math Solver with interface to input a question and steps involved to get the solution. We only solved elementary math word problems involving basic math operations like addition, subtraction, multiplication and division. The solver can be used for tutoring school children the basic math word problems , this also can be used as a guide to understand sentences containing mathematical quantities.

We are still a long way from modeling how humans solve word problems. But, with the advancement in the area of deep learning , there has been renewed interest in this area recently. Natural Language Understanding is a vital step for solving word problems. There is also need for inclusion of world knowledge to solve specific problems. For example:- ``Robert earns 5 dollars per day, how much does he earn in a week?'', the answer to this problem requires the knowledge of the relation between week and day. Solving these kinds of problems which are popularly known as unit conversion problems, requires knowledge about different units and their significance.

Our main motivation comes from the fact that many children tend to repeat mistakes in arithmetic word problems because they fail to see a relationship between the problem and the corresponding mathematical representation or equation. Solving a mathematical equation is easier than mapping to a mathematical formula. In this work, we intend to bridge this gap where each sentence in a problem can be figuratively shown their role in the final equation generation.

In this paper, we represent the word problem as a graph consisting of nodes. Each node contains information of the entity, quantity of the entity, the entity holder, the beneficiary if any transfer is done, the time, place or other information related to the entity. The nodes are called frames. A frame can either be a state frame or an action frame. The creation of a frame is usually triggered by the verbs present inside a sentence. State frames refer to some world state while the action frames act on state frames either to modify the quantities of different entities involved in particular state frames or to create new state frames. The frames which act on same units or entities can only interact among themselves. For any arithmetic operation, two quantities should have the same unit. In our framework, we take this fact into account. The main motivation was to make the solving process of arithmetic problem simpler. The frame wise representation of the problem makes it easier for a kid to understand the word problem, the related quantities and interplay between them. This whole process can be simulated with diagrams to make understanding of the word problems much clearer and less complicated.
\section{Previous Work}
\label{prev}
Yefim Bakman \cite{bakman2007robust} touched upon the understanding involving word problems with extraneous information. Mukherjee and Garain \cite{mukherjee2008review} surveyed on different techniques used for word problem solving. Word problems can be represented as a relation model as done by Bobrow \cite{bobrow1964question}. Multiple Approaches were tried for reasoning across sentences of the problem text. Approaches can be template alignment, prediction of verb categories and solving the problem, using CFG rules along with the generation of equation trees. The system designed by Kushman et al \cite{kushman2014learning} was a joint log linear distribution over the full set of equations and alignments between the variables and text. The number of equations was determined by the number of training equation templates. The number slots were filled by the numbers present in the text while the unknown or variable slot were filled by the nouns in the problem text. The derivations possible out of the all templates, the derivation with the highest probability score was chosen. Illinois Math Solver \cite{roy2016illinois} had two modules to solve any arithmetic word problem. First module was a CFG based Semantic Parser that would solve simple arithmetic problems where specific keywords like add, subtract, difference, product, sum, multiply or divide were present. If the problem could not be solved by the first module, the problem was solved by decomposition of the arithmetic problem into a series of classification problems \cite{roy2016solving}. The classification outputs were then combined to form an expression tree through constrained inference. Hosseini's \cite{hosseini2014learning} ARIS system used verb categorization. ARIS analyzed the sentences in the problem for identifying relevant variables, their values and mapping this information into a set of linear equations which can be easily solved. The system identified 7 kinds of verbs used in the problems which was predicted by support vector machines. Mitra and Baral \cite{mitra2016learning} created an arithmetic word problem solver which learned how to use formulas to solve simple addition and subtraction problems. The formulas were modeled as templates with pre-defined slots. They used a log-linear model to find out the best possible formula to solve a problem. The features to the model were dependency labels by running Stanford dependency parser, POS tags, some linguistic cues, Wordnet \cite{miller1995wordnet} features.
\section{Corpus Annotation}
\label{cor}
The main task in this problem is the identification of the frames. We needed to create a corpus of frame annotated sentences to create automatic frame identification module.
\begin{table}[htbp]
\centering
\caption{Annotated Corpus Size}
\begin{tabular}{cccc}\hline
\textbf{Type} & {\textbf{\#Questions}} & \textbf{\#Sentences } & \textbf{\#Frames}\\ 
\hline
Train & 444 & 1253 & 1253\\
Test & 60 & 168 & 168\\
\textbf{Total} & 504 & 1421 & 1421\\\hline
\end{tabular}
\label{corpus}
\end{table}
Each sentence in our approach contained a frame information. We created an off-line tool using python programming language for facilitating annotation. The number of frame types were fixed to 22. Words with similar meaning evoke a frame. The similar words can be found out from the frame-net. We created a list of frames and a list of words corresponding to a frame. The kappa score \footnote{https://en.wikipedia.org/wiki/Fleiss's\_kappa}The inter-annotator's agreement for frame annotation was 0.85 which is considered perfect agreement. Two annotators participated in the annotation task.
\section{Approach}
\label{approach}
Our approach focuses more on extraction of knowledge from sentences, parse information of the sentences, construct frames and finally solve the problem. Our approach is similar to Sundaram and Khemani \cite{sundaram2015natural}. We modeled every word problem as a graph of state and action frames. Action frames act on their respective state frames and state frames undergo change in quantity. The intended question in a word problem is about specific slots in frames, most of the times the quantity which are explained in the subsection \ref{aps}. Our approach has the following steps:
\begin{itemize}
	\item Preprocessing: Conversion of numbers in words to actual numbers
	\item Identification of Frames
    \item Parse Sentences
    \item Fill slots corresponding to frames
    \item Build a graph of frames
    \item Traverse the graph to find the answer
\end{itemize}
\begin{figure}[ht]
\caption{Example of state and action frames interaction}
\centering
\includegraphics[width=0.5\textwidth]{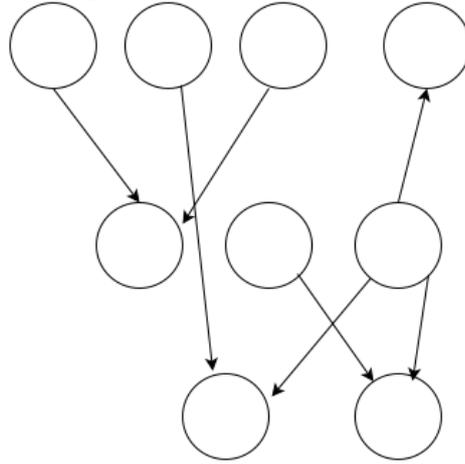}
\end{figure}
\subsection{Approaches for Frame Identification}
Frame Identification constituted the first step in our approach. We implemented different machine learning approaches for creating frame identifier. The design of the classifiers were implemented using sklearn \cite{Pedregosa:2011:SML:1953048.2078195} machine learning library. The classifiers used were : Support Vector Machines \cite{cortes1995support} and Random Forests \cite{breiman2001random}. We did not use any neural network approaches because of limited number of training examples.
\subsubsection{TF-IDF Features}
TF-IDF \cite{sparck1972statistical} assigns weights to words (or n-grams) based on its frequency in a document and its frequencies across documents to find out how important a word is to a document in a corpus. In this case, a document is a sentence. TF-IDF was calculated for word unigrams (uni), bigrams (uni-bi). We also experimented with character n-grams in different ranges [2-6 and 3-6]. We did not use any additional lexical or linguistic features like parts-of-speech tags, morph features, wordnet \cite{miller1995wordnet} features or dependency labels for frame identification.
\subsection{Approach for Problem Solving}
\label{aps}
After the frame identification, we parsed each sentence to fill the frame slots. For dependency parsing, Stanford dependency parser \cite{chen2014fast} which was a neural network based parser was used. We used following slots or attributes for each frame which were identified from specific dependency labels whose dependency mapping are given below.
\begin{table}{ht}
\label{mapping}
\caption{Dependency label to Frame Slot Mapping}
\centering
\begin{tabular}{cc}\hline
\textbf{Dependency~Label} & \textbf{Frame Slot}\\\hline
Subject & Entity Holder\\
dobj    & Entity\\
amod    & Attribute of Entity\\
iobj    & Beneficiary\\
nummod  & Quantity\\
nmod:case & Addidional Info\\
\hline

\end{tabular}
\end{table}
After each frame was created, a graph was built with frames. Action frames acted on a set of state frames to force a change in quantities contained in frames. The quantity updates could be intimated to neighboring frames so that all the frames were updated at once. Each question queried about specific frame slots which could be easily answered by traversing the graph once. 
\section{Working Example of the System}
For the arithmetic word problem: ``John had 5 books. John gave Robert 2 books. How many books John have now?''
\begin{figure}
  \includegraphics[scale=0.6,center]{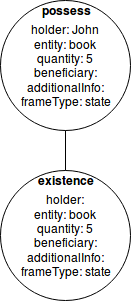}\\
  \caption{Initial Frames} \label{fig:2}
\end{figure}
the equation for this question is $x = 5 - 2$ and the solution is $x = 5$ 
\begin{figure}
  \includegraphics[scale=0.7]{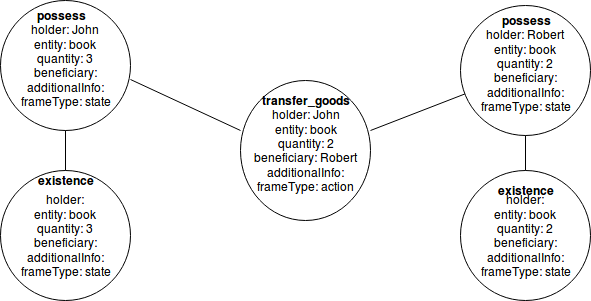}\\
  \caption{Complete Graph of Frames} \label{fig:3}
\end{figure}
Figure \ref{fig:2} shows the initial frames created after parsing the first sentence. In word problems, sometimes questions are asked on information which are not present explicitly in the question. The answer to this kind of question can only be answered through proper inference. If the question in the above example is changed to ``How many books are there?'', a solver needs to infer that ``Somebody has some books'' means ``There are some books.'' We tried to incorporate this information in our system. So in figure \ref{fig:2}, there are two frames connected to each other instead of one. Both these frames are state frames. The second sentence gives the information about a transfer operation carried out between two entity holders. The order and type of operation can be found out by matching the entity holders and entities. So in this case, the transfer\_goods frame triggers a subtraction operation in one possess frame with an update in quantity slot $5 - 2 = 3$. It also creates another possess frame triggering an addition operation with quantity coming from $0 + 2 = 2$. As the graph is built dynamically with addition of new information, the updated quantities information are propagated to all the connected nodes. The question sentence is also parsed to find out the frame type and type of question asked. ``who'' kind of question seeks answer from entity holder slots of the frames, similarly ``what'' maps to entity slots. ``how many'' questions interrogates about the quantities involved in the frames. In the current system, the relation between all these question types and frame slots are predefined.
\section{Experimental Results}

Table \textbf{3} illustrates the results of the same for the aforementioned classifiers. The best performing metrics are shown in bold. Table \textbf{4} shows the comparison of our system with ARIS. Our system was able to solve 118 questions out of 312 questions of the AI2 dataset. \footnote{http://ai2-website.s3.amazonaws.com/data/arithmeticquestions.pdf} 
\begin{table}[ht]
\label{frame}
\caption{Frame Classification with TF-IDF Features}. 
\centering
\begin{tabular}{c| c |c |c |c}
\textbf{Model} &  \textbf{Features}  & \textbf{Precision} & \textbf{Recall} & \textbf{F1-Score} \\ \hline
&uni & \textbf{0.82}   &   \textbf{0.84}   &   \textbf{0.81}\\
Linear-SVM & uni-bi & 0.75   &   0.81   &   0.77\\
&char[2,6] & 0.73   &   0.80   &   0.75\\
&char[3,6] & 0.75   &   0.81   &   0.76\\\hline
&uni & 0.79   &   0.82   &   0.78\\
Random-Forest & uni-bi & 0.74   &   0.76   &   0.73\\
&char[2,6] & 0.75   &   0.80   &   0.76\\
&char[3,6] & 0.77   &   0.81   &   0.78\\
\hline

\end{tabular}
\end{table}
\begin{table}{ht}
\label{Results}
\caption{Comparison of System Accuracy with ARIS}
\centering
\begin{tabular}{cc}
\textbf{System} & \textbf{Accuracy}\\\hline
ARIS & 77.7\%\\
Our System & 37.8\%\\
\end{tabular}
\end{table}
\section{Error Analysis and Observation}
One source of error could be incorrect frame identification. The major limitation of our system is its dependence on dependency parser's output. The example below shows how parsing errors can propagated to result in an incorrect answer or unsolvable problem.
\begin{itemize}
	\item ``How many Pokemon cards does Jason have now ?''
    \item The parser wrongly tags ``now'' as the entity which results in the algorithm not being able to find the answer from the graph.
\end{itemize}
If the parser performs poorly, then the slots will be incorrectly populated. This will finally impact the overall performance in solving the problem. Unit conversion problems can not be handled by the current system e.g ``How many days are there in a week?''. Similarly questions like ``Find the answer when 3 is added to 5.'' can not be solved because the problem has a different kind of structure to basic word problems and the dependency labels given by the parser to not conform to our mapping. We observed that the linear-SVM outperformed random forests. We also observed that the word unigram (uni) TF-IDF features were better in classification of frames than character n-gram TF-IDF features. Almost all the arithmetic problem solvers output only the answers or equation, but our system outputs step wise explanation along with the answer and equation.

Even though the motivation  of the design of the frames came from framenet \cite{baker1998berkeley}, the output of our system is not similar to framenets. If a verb had different meanings, we did not create different frames for different semantics. We focused more on the computational part of the involved frames.
\section{Conclusion and Future Work}
In this paper, we present an easily understandable framework for solving arithmetic word problems. We hope that this will motivate to try out different techniques to help students understand word problem solving better.
In our approach, we have predefined action frames performing arithmetic operations. This task can be learned which action frame does what operation. The frame identification module can be improved by increasing its accuracy. We intent to increase the training data size by bootstrapping \cite{dietterich2000ensemble}. Word embeddings can be used to find similar verbs. We rely on parser's output to fill the frame slots, we could explore the use of computation parsing instead of linguistic parsing. We could also integrate a co-reference and anaphora resolution module into our system as it is not handled in the current set-up. The learning of the question types and slots can also explored.
\label{confu}

\bibliography{llncs}
\bibliographystyle{splncs03}
\end{document}